\documentclass[letterpaper, 10 pt, conference]{IEEEconf}  

\IEEEoverridecommandlockouts                              

\usepackage{amsmath,amssymb}
\usepackage{algorithm}
\usepackage{algorithmic}
\usepackage{graphicx}
\usepackage{booktabs}
\usepackage{enumerate}

\usepackage{float} 
\usepackage{xcolor}
\usepackage{listings}
\usepackage{subcaption}
\usepackage{multirow}
\usepackage{listings}
\lstdefinestyle{json}{
	basicstyle=\ttfamily\tiny,         
	breaklines=true,                   
	columns=flexible,                  
	frame=tb,                          
	showstringspaces=false,            
	stringstyle=\color{blue},          
	keywordstyle=\color{red},          
	commentstyle=\color{green!60!black}, 
	literate={á}{{\'a}}1 {ñ}{{\~n}}1, 
}

\DeclareMathOperator*{\argmin}{arg\,min}

\title{\LARGE \bf
Perceiving, Reasoning, Adapting: A Dual-Layer Framework for VLM-Guided Precision Robotic Manipulation
}

\author{Qingxuan Jia$^{1}$, Guoqin Tang, Zeyuan Huang, Zixuan Hao, Ning Ji, Shihang, Yin, Gang Chen$^{*}$%
\thanks{*This work was supported by National Natural Science Foundation of China (No.62173044); Major Project of the New Generation of Artificial Intelligence of China (No.2018AAA0102904); and State Key Laboratory of Robotics and Systems (HIT) (SKLRS-2025-KF-07). All authors have read and agreed to the published version of the manuscript.}
}
\begin{document}

\maketitle
\thispagestyle{empty}
\pagestyle{empty}

\begin{abstract}
Vision-Language Models (VLMs) demonstrate remarkable potential in robotic manipulation, yet challenges persist in executing complex fine manipulation tasks with high speed and precision. While excelling at high-level planning, existing VLM methods struggle to guide robots through precise sequences of fine motor actions. To address this limitation, we introduce a progressive VLM planning algorithm that empowers robots to perform fast, precise, and error-correctable fine manipulation. Our method decomposes complex tasks into sub-actions and maintains three key data structures: task memory structure, 2D topology graphs, and 3D spatial networks, achieving high-precision spatial-semantic fusion. These three components collectively accumulate and store critical information throughout task execution, providing rich context for our task-oriented VLM interaction mechanism. This enables VLMs to dynamically adjust guidance based on real-time feedback, generating precise action plans and facilitating step-wise error correction. Experimental validation on complex assembly tasks demonstrates that our algorithm effectively guides robots to rapidly and precisely accomplish fine manipulation in challenging scenarios, significantly advancing robot intelligence for precision tasks.
\end{abstract}

\section{INTRODUCTION}

Robotic assembly tasks are crucial in modern manufacturing, involving precise manipulations in environments with complex spatial relationships. These tasks, ranging from electronics assembly to automotive production, require robots to understand intricate spatial configurations and execute precise movements, making autonomous execution challenging. Achieving reliable autonomous assembly could significantly enhance efficiency, reduce costs, and improve safety in industrial settings \cite{r1-smartmanu}.

The key challenges in robotic assembly include understanding complex spatial relationships between objects; the need for both coarse motion planning for approach and fine control for precise insertion; and the requirement for maintaining spatial-semantic coherence throughout task execution. Traditional robotic assembly systems rely on a pipeline of perception, task planning, and execution control. While methods using geometric planning \cite{r2,r3} and pure vision-based approaches \cite{r4} have achieved success in structured settings, they struggle with complex spatial reasoning tasks. These systems typically require extensive manual programming for each new task, lack integrated spatial-semantic understanding, and have limited ability to recover from errors during fine manipulation\cite{r10-weakness}.

The advent of vision-language models (VLMs) has opened new avenues for more flexible and adaptive task planning. VLMs, capable of interpreting both visual scenes and natural language, offer potential for overcoming the limitations of traditional systems. Recent studies have explored integrating VLMs with robotic control to improve task flexibility \cite{r6-vlm1,r7-vlm2,r5-vlm0}. However, existing models still face significant challenges: while VLMs excel at high-level semantic understanding and task planning, they struggle with precise spatial reasoning needed for fine manipulation tasks. Most VLM-based robotic systems lack efficient mechanisms to represent and reason about complex 3D spatial relationships, limiting their effectiveness in precision assembly tasks \cite{r8-lack3d,r9-leak3d,r11-lack3d}.

This paper presents a novel progressive VLM planning algorithm for robotic assembly, bridging the gap between high-level semantic understanding and precise spatial control. Our approach demonstrates significant improvements in success rates, reduced planning time, and better precision in complex assembly scenarios compared to traditional methods. We maintain real-time performance while achieving millimeter-level precision in final assembly. Our key innovations include:

\begin{itemize}
	\item A dual-layer spatial-semantic fusion framework enabling VLMs to understand complex spatial configurations without 3D-specific training.
	\item A spatio-temporal memory framework for tracking object positions, semantics, and task progress in complex environments.
	\item A progressive VLM interaction strategy that adjusts robot guidance based on environment, experience, and task history.
\end{itemize}

\begin{figure*}[ht] 
	\centering
	\includegraphics[width=\textwidth]{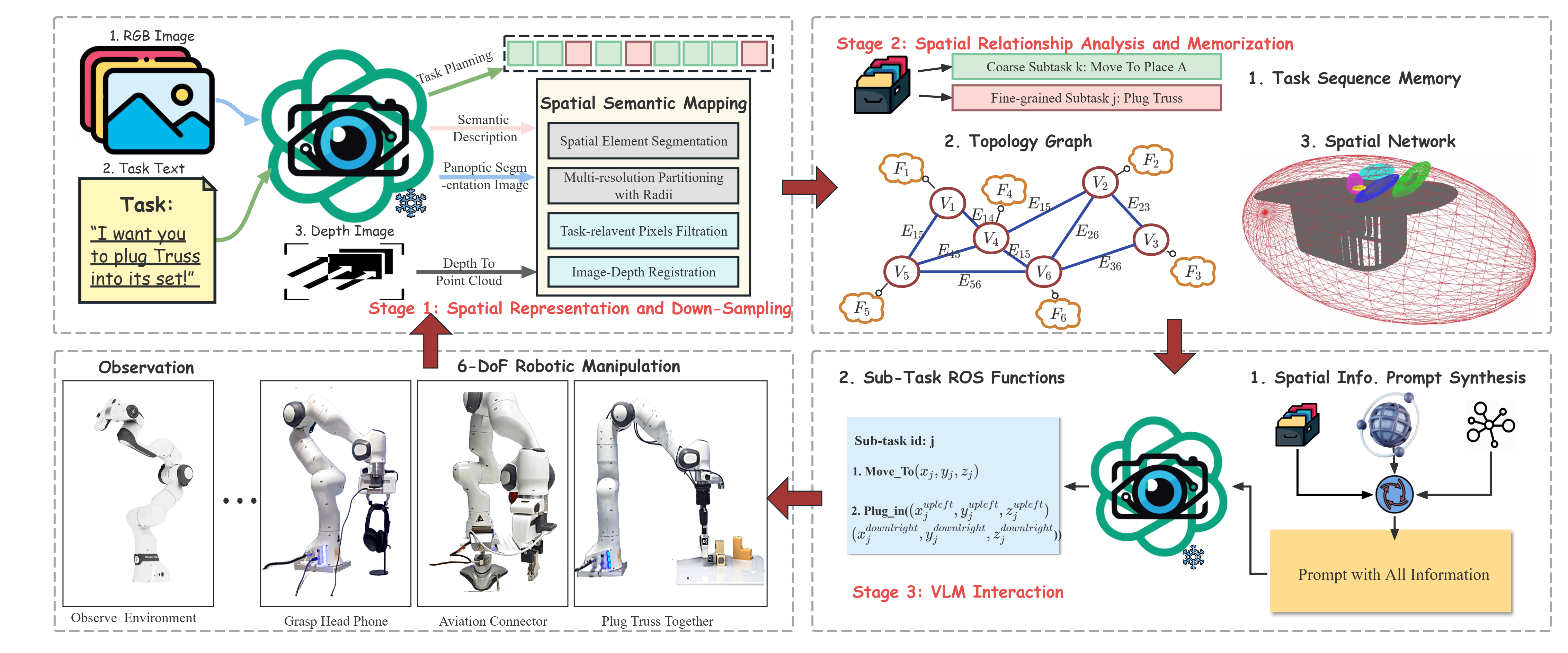}
	
	\caption{Progressive VLM Planning Algorithm Architecture. \textbf{Top-Left:} Stage 1 - Spatial Representation and Down-sampling.  Stage 1 takes RGB-D images and task descriptions as input. It performs initial VLM interaction for panoptic segmentation, registering it with depth point clouds to obtain spatial representations. \textbf{Top-Right:} Stage 2 - Spatial Relationship Analysis and Memorization. Stage 2 constructs and maintains task sequence memory, topology graph, and spatial network based on registered point clouds for spatiotemporal scene understanding. \textbf{Bottom-Right:} Stage 3 - Task-Oriented VLM Interaction. Stage 3 synthesizes VLM prompts using task memory, topology graph, and spatial network information. These prompts guide the VLM to generate ROS functions for robot subtask execution. \textbf{Bottom-Left:} Robot Execution. A real-world image depicting the robot arm executing a subtask based on the ROS functions. \textbf{Feedback Loop:} Feedback Loop: After each subtask, the robot re-observes the environment to obtain scene information and task status, providing input for the next iteration in a closed-loop system.} 
	
	\label{fig:architecture}
\end{figure*}

\section{RELATED WORKS}
\subsection{Spatial Reasoning in Large Language Model}
Recent works in large language model-based spatial reasoning primarily follow two approaches. Multi-modal fusion methods, as demonstrated in \cite{r12-Palm-E} and \cite{r13-PaLi-x}, integrate visual embeddings with text inputs for comprehensive scene understanding. While effective, these approaches demand extensive training data and computational resources. The alternative strategy employs frozen models with prompts. Works like \cite{r14-clipFO3D, r16-rtgrasp} and \cite{r15-rt2} combine language guidance with visual-language models, enabling generalization to novel environments. However, these methods often struggle with precise spatial relationships and complex manipulation tasks, particularly in dynamic settings.

Both approaches have limitations: spatial reasoning is limited by language model abstraction, high computational costs restrict scalability, and existing methods struggle with long-term spatiotemporal reasoning.

\subsection{Large Language Model Applications in Robotic Precision Tasks}
Large language models in robotic precision tasks fall into three distinct categories. Task-specific reasoning models, exemplified by \cite{r17-copal}, employ specialized architectures with multi-layer feedback mechanisms that enable dynamic plan correction but lack cross-domain generalization capabilities. End-to-end trained models such as \cite{r18-scmml} and \cite{r19-pa} connect perception directly to control outputs, simplifying system design at the cost of interpretability and vulnerability to data distribution biases. Strategic interaction frameworks, represented by \cite{r20} and \cite{r21}, utilize frozen pretrained models with carefully designed interaction strategies, offering better adaptation to diverse task descriptions but with performance heavily dependent on algorithm design rather than fully leveraging model capabilities.

These approaches share a fundamental limitation in maintaining spatio-temporal relationships. They lack mechanisms for continuously tracking spatial changes while integrating semantic understanding with precise geometric information, highlighting the need for our approach of maintaining both topological and spatial representations throughout task execution.

\section{Methods}
Our method addresses the challenges of dynamic robotic manipulation through a novel integration of visual language models and progressive planning. Whole structure see\ref{fig:architecture}. Before detailing our approach, we first establish the key mathematical foundations and notations.
\subsection{Preliminaries and Notations}
To facilitate the technical discussion of our method, we define the essential parameters and mathematical notations that will be used throughout this paper:

\begin{itemize}
	\item RGB-D camera observations at timestamp $t$:
	\begin{equation}
		z_t = \{I_t, D_t\} \in \mathbb{R}^{H \times W \times (3+1)}
	\end{equation}
	where $I_t$ is the RGB image and $D_t$ is the depth map.
	
	\item Scene representation parameters:
	\begin{equation}
		\begin{aligned}
			G_t &= (V_t, E_t, F_t) &: \text{topology graph} \\
			S_t &= \{s_{ij,t}\} &: \text{spatial network} \\
			\mathcal{P}_t &= \{p_k^{(i)}\}_{i=1}^N &: \text{pixel set}
		\end{aligned}
	\end{equation}

	\item Task Memory Structure: 
	\begin{equation}
		\begin{aligned}
			\mathcal{M} = \{TTP&,\, SS,\, MSH\},\\
			TTP &:\; \text{Task Topology Path} \\
			SS &:\; \text{Subtask Status}\\
			MSH &:\; \text{Motion Sequence History}
		\end{aligned}
	\end{equation}
\end{itemize}

\subsection{Progressive VLM Planning Algorithm}

 To handling Traditional robotic manipulation approaches separate planning from execution, leading to reduced adaptability in dynamic environments. We propose a self-supervised progressive planning framework that enables real-time adaptation through continuous VLM interaction. Our key innovation lies in the tight integration of perception and action, where VLM continuously guides both scene understanding and manipulation strategy refinement. The system adaptively switches between two execution modes based on the robot's proximity to manipulation targets:
\begin{itemize}
	\item \textbf{Coarse motion ($distance > \tau$):} Prioritizes efficient global navigation while maintaining safety
	\item \textbf{Fine manipulation ($distance \leq\tau$):} Focuses computational resources on precise control and error recovery
\end{itemize}
Our framework implements this adaptive behavior through three interconnected stages, each feeding information to the next while receiving feedback for continuous refinement:

\subsubsection{Stage 1 - Spatial-Semantic Mapping}

The first stage creates a foundational spatial representation by combining sphere-based partitioning with VLM-guided semantic filtering. This stage ensures efficient computation by prioritizing precision near the robot while maintaining a broad understanding of the environment. Further details on the spatial-semantic mapping method are presented in Section \ref{space_seg_reg}.

\subsubsection{Stage 2 - Scene Understanding}

Building on the spatial-semantic mapping, this stage maintains a topological graph for semantic relationships and a spatial network for detailed geometric information. The specifics of the scene understanding mechanism are presented in Section \ref{recons_maintain}.

\subsubsection{Stage 3 - Task-Oriented VLM Interaction}
In this final stage, we focus on how VLM interacts with the system to generate and refine action plans. Based on the robot’s proximity to the manipulation target, the system operates in two distinct modes:

\paragraph{Coarse Motion} During coarse motion, VLM generates high-level prompts based on global relationships and constraints. The system focuses on optimizing path efficiency while avoiding obstacles and considering environmental changes. Outputs include motion strategies and safety parameters.

\paragraph{Fine Manipulation} As the end-effector approaches the target, the system transitions into fine manipulation mode, concentrating resources on precision control and error recovery. In this mode, VLM monitors execution progress and dynamically adjusts control parameters based on real-time feedback. Outputs include precise control parameters and failure recovery strategies. Details see Section \ref{task_oriented_vlm_interaction}.

\subsubsection{Algorithm and System Execution Flow}

To formalize this process, Algorithm \ref{alg:progressive_planning} demonstrates the step-wise integration of these stages into a cohesive execution framework. The algorithm highlights how the system cycles through stages 1 and 2, transitioning between coarse and fine motion strategies based on task progression.

\begin{algorithm}[ht]
	\caption{Progressive VLM Planning Algorithm\label{alg:progressive_planning}}
	\begin{algorithmic}[1]
		\REQUIRE RGB-D image $I$, camera parameters $K$, task description $T$
		\ENSURE Task execution status and feedback
		\STATE Initialize max\_iterations, $\tau=r_1$, iteration\_count $\leftarrow 0$
		\STATE Initialize task memory $\mathcal{M}$ with task topology path $TTP$, subtask status $SS$, and motion sequence history $MSH$
		\WHILE{task not complete \&\& iteration\_count $\leq$ max\_iterations}
		\STATE // Stage 1: Build spatial-semantic representation
		\STATE Segment and filter image by VLM
		\STATE Partition space elements
		\STATE Establish spatial mapping
		
		\STATE // Stage 2: Maintain scene understanding
		\STATE Initialize/update 2D semantic topology graph $G = (V, E, F)$
		\STATE Initialize/update 3D spatial network $S$
		
		\STATE // Stage 3: Execute appropriate motion strategy
		\IF{$\|\text{target} - \text{end\_effector}\| > \tau$}
		\STATE Generate coarse motion prompt $Prompt_{coarse}$ using $TTP$, $E$, $SS$, $S$, and $MSH$
		\STATE Execute coarse motion planning based on VLM response
		\ELSE
		\WHILE{fine manipulation not successful}
		\STATE Generate fine manipulation prompt $Prompt_{fine}$ using $TTP$, $S_{local}$, sensor data $D_t$, $SS$, and $MSH$
		\STATE Execute precision-guided manipulation based on VLM response
		\STATE Adapt based on real-time feedback
		\ENDWHILE
		\ENDIF
		
		\STATE Update execution results $R_t$ and scene state $S_t$
		\STATE Update task memory: $\mathcal{M}_{t+1} = \text{Update}(\mathcal{M}_t, R_t, S_t)$
		\STATE iteration\_count $\leftarrow$ iteration\_count + 1
		\ENDWHILE
	\end{algorithmic}
\end{algorithm}

\subsection{Space Element Segmentation and Pixel Registration Strategy}
\label{space_seg_reg}

\begin{figure}[t]
	\centering
	\begin{subfigure}[b]{0.45\linewidth}
		\centering
		\includegraphics[width=\linewidth]{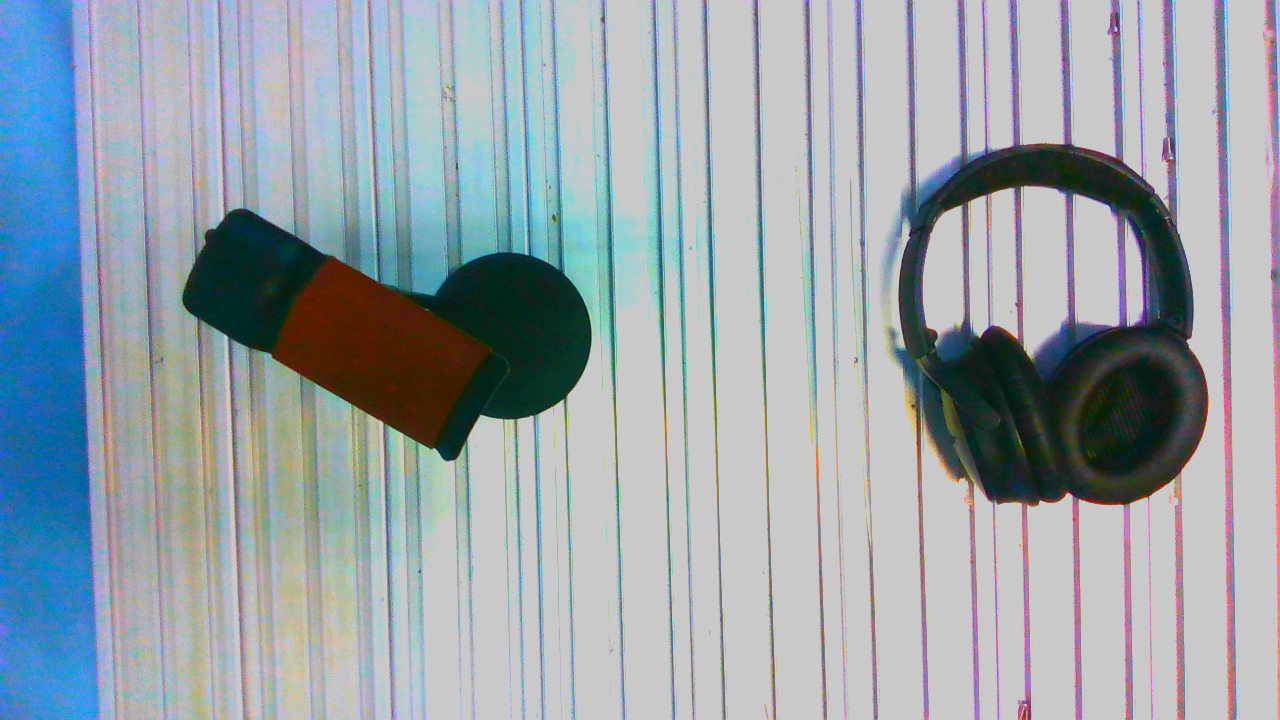}
		\caption{Original RGB image}
		\label{fig:original_rgb}
	\end{subfigure}
	\hfill
	\begin{subfigure}[b]{0.45\linewidth}
		\centering
		\includegraphics[width=\linewidth]{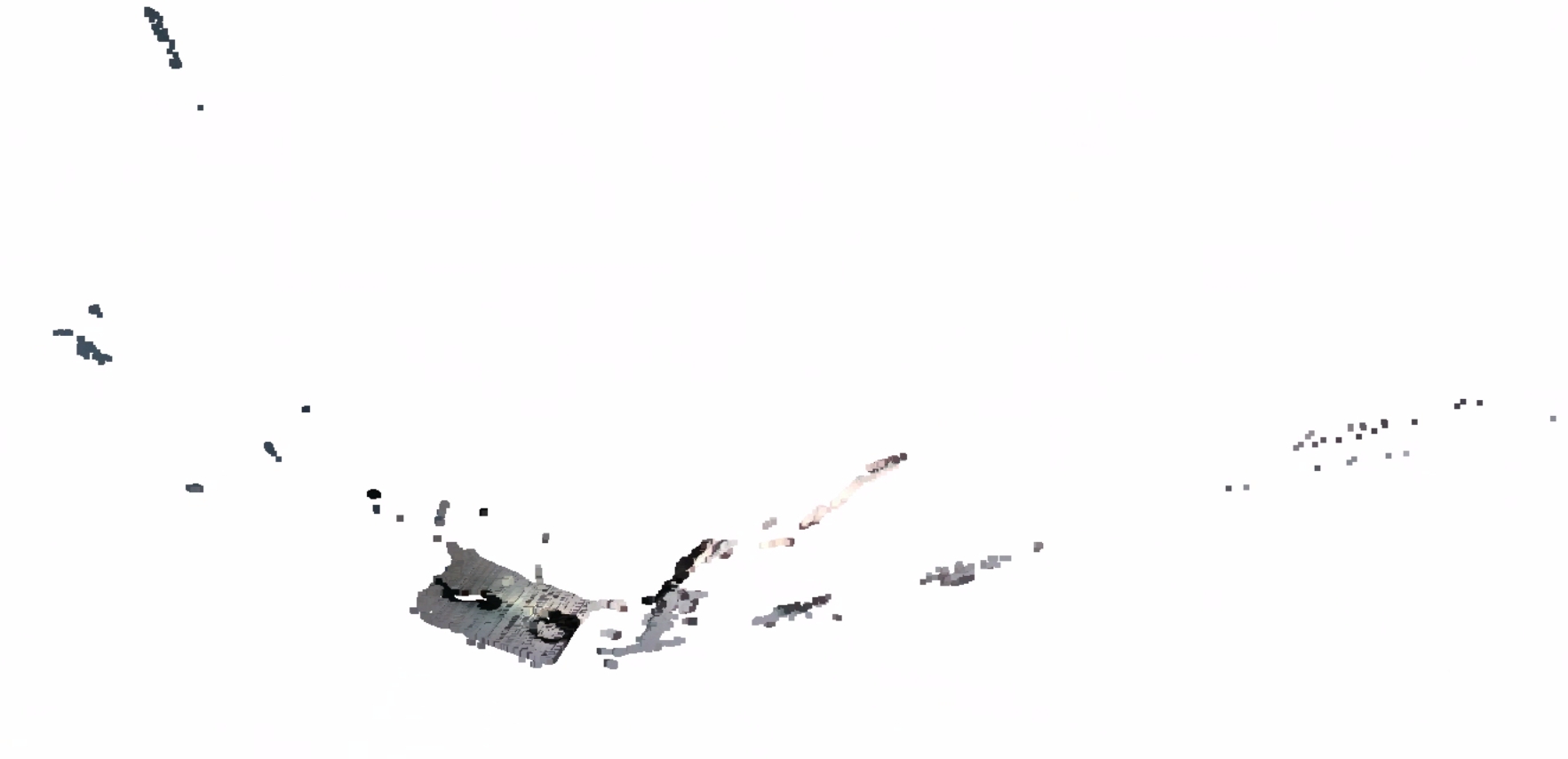}
		\caption{Colored point cloud}
		\label{fig:colored_point_cloud}
	\end{subfigure}
	
	\vspace{0.5em}
	
	\begin{subfigure}[b]{0.45\linewidth}
		\centering
		\includegraphics[width=\linewidth]{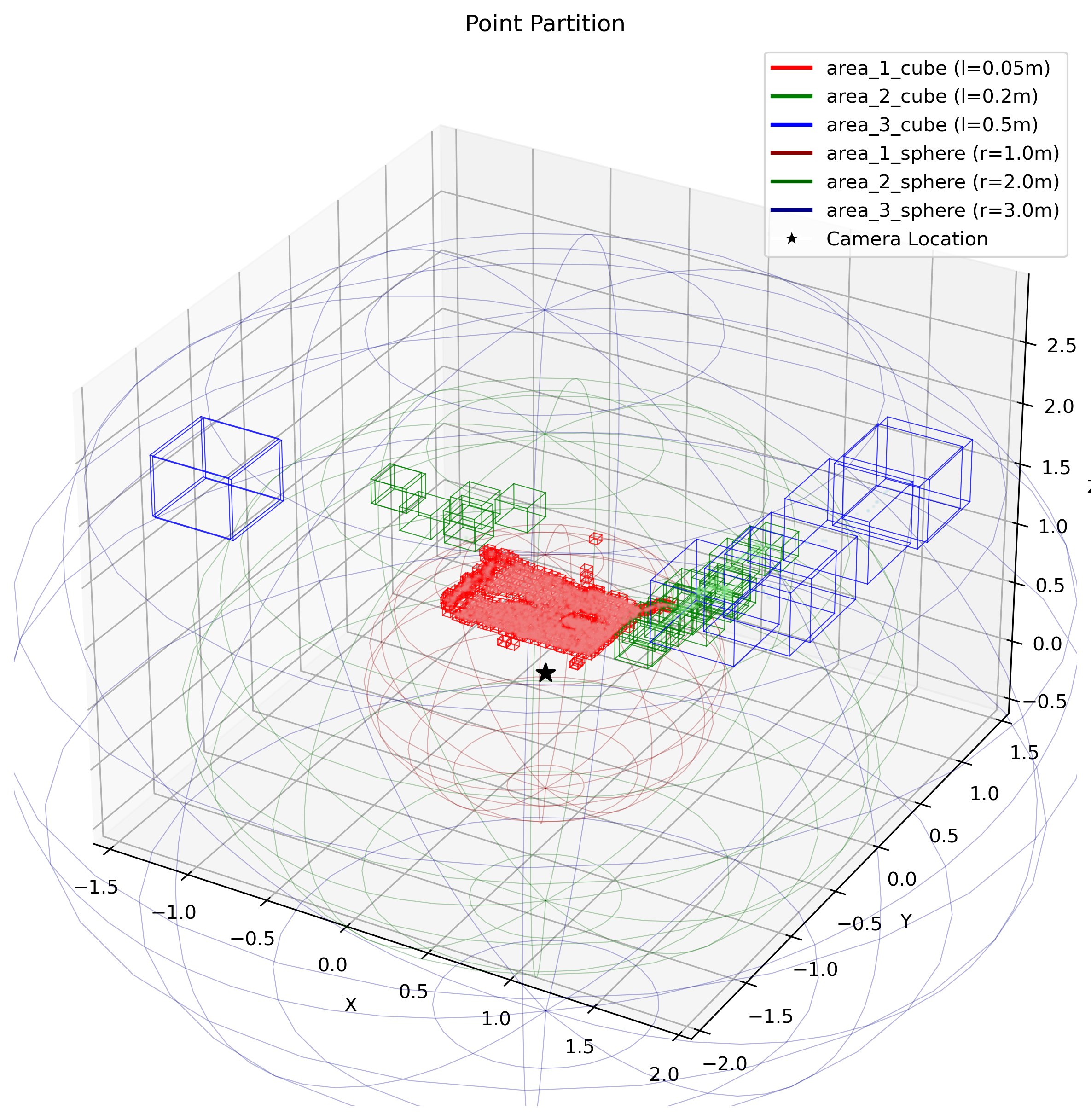}
		\caption{Spatial sphere segmentation}
		\label{fig:spatial_segmentation}
	\end{subfigure}
	\hfill
	\begin{subfigure}[b]{0.45\linewidth}
		\centering
		\includegraphics[width=\linewidth,trim={500 100 500 100},clip]{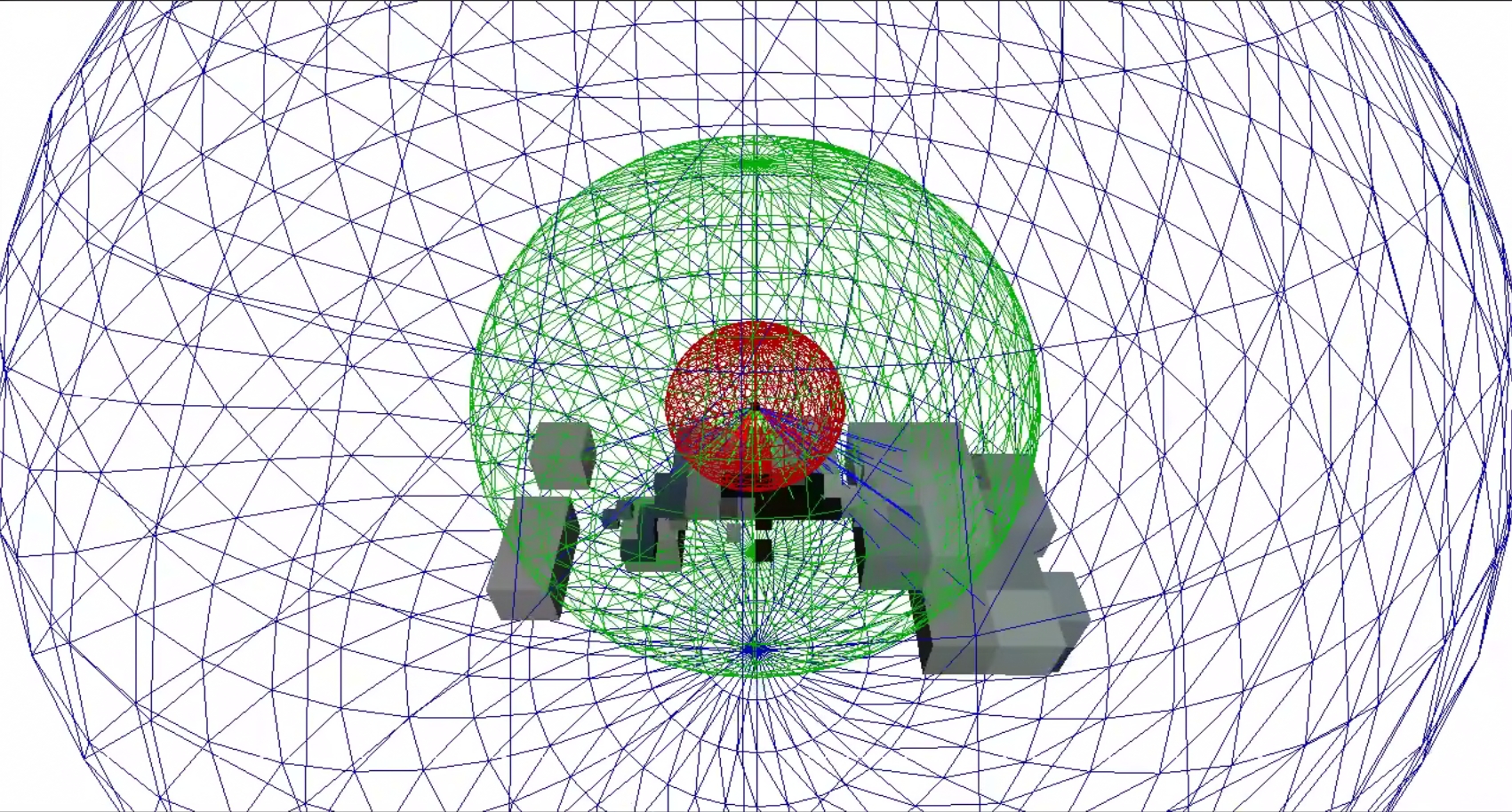}
		\caption{Dual representation mapping}
		\label{fig:mapping_relationships}
	\end{subfigure}
	
	\caption{Segmentation and Mapping}
	\label{fig:progressive_mapping}
\end{figure}

Effective robotic manipulation requires precise spatial understanding near the manipulation target, while maintaining efficient processing of the broader environment. We address this challenge through a novel multi-resolution approach that adaptively manages computational resources based on task requirements and spatial proximity. See \ref{fig:spatial_segmentation}

\subsubsection{Universal Spatial Representation}
The first step is establishing a precise mapping between camera observations and the robot's operational space. Given a depth image \( D_t \in \mathbb{R}^{H \times W} \) and the camera intrinsic matrix \( K \), we compute universal spatial coordinates via a two-stage transformation process.

We transform pixel observations to the robot's universal frame through the following chain of transformations:
\begin{multline}
	\mathbf{p}_b = 
	\underbrace{
		\begin{bmatrix}
			{}^b\mathbf{R}_c & {}^b\mathbf{t}_c \\
			\mathbf{0}^T & 1
		\end{bmatrix}
	}_{{}^{b}\mathbf{T}_c}
	\cdot
	\left( 
	d K^{-1} \begin{bmatrix} u \\ v \\ 1 \end{bmatrix} 
	\right) \\
	\text{where } {}^{b}\mathbf{T}_c = {}^{b}\mathbf{T}_e \cdot {}^{e}\mathbf{T}_c
	\label{eq:compact_transform}
\end{multline}
The transformation matrix \( {}^b\mathbf{T}_c \) is the product of:
\begin{itemize}
	\item \( {}^b\mathbf{T}_e \): End-effector to base frame (from robot kinematics)
	\item \( {}^e\mathbf{T}_c \): Camera to end-effector frame (calibration matrix)
\end{itemize}
This projection enables precise mapping from image pixels to the robot's operational space.

\subsubsection{Multi-Resolution Space Partitioning}
We partition the operational space into three concentric zones, employing an octree-inspired discretization approach. The space is characterized by two distance thresholds:
\begin{equation}
	\begin{aligned}
		r_1 &: \text{threshold for high-resolution processing} \\
		r_2 &: \text{boundary of mid-resolution zone}
	\end{aligned}
\end{equation}

For efficient spatial representation, we discretize each zone using cubic elements of different edge lengths (\(l_1 < l_2 < l_3\)). The spatial elements $\mathcal{E}$ in each zone are defined as:
\begin{equation}
	\mathcal{E}(\mathbf{p}) = \begin{cases}
		\text{CenterOf}(\text{Cube}(l_1)), & \|\mathbf{p}\| < r_1 \\
		\text{CenterOf}(\text{Cube}(l_2)), & r_1 \leq \|\mathbf{p}\| < r_2 \\
		\text{CenterOf}(\text{Cube}(l_3)), & \|\mathbf{p}\| \geq r_2
	\end{cases}
\end{equation}

This multi-resolution scheme ensures high precision in the manipulation zone while maintaining computational efficiency in farther regions. When processing point cloud data, each point is mapped to its corresponding cubic element's center position, providing an effective balance between spatial accuracy and computational cost.

\subsubsection{Task-Relevant Pixel Selection}
The VLM performs panoptic segmentation on the RGB image to identify task-relevant objects, where each pixel is assigned a semantic label \( l \) and instance ID \( i \):

\begin{equation}
	\mathcal{P}_{task} = \{(u,v) \mid \text{VLM}(I_t, \mathcal{T}) \}
\end{equation}
where \( T \) is the task description.

\subsubsection{Ray-based Registration and Denoising}
For each task-relevant pixel \( (u,v) \), we cast a ray from the camera center through the pixel:

\begin{equation}
	\mathbf{r}(t) = t K^{-1} \begin{bmatrix} u \\ v \\ 1 \end{bmatrix}, \quad t > 0
\end{equation}

This ray intersects multiple space elements. We establish initial mappings between pixels and space elements through these intersections:

\begin{equation}
	\mathcal{Map}(u,v) = \{\mathcal{E}_i^k \mid \mathbf{r}(t) \in \mathcal{E}_i^k \text{ for some } t > 0\}
\end{equation}

The resulting mapping contributes to both the topology graph and spatial network in Stage 2 and the subsequent task planning in Stage 3.

\subsection{Spatial Relationship Reconstruction and Topology Graph Maintenance}
\label{recons_maintain}
\begin{figure}[htbp]
	\centering
	\begin{minipage}[b]{0.20\textwidth}
		\centering
		\includegraphics[width=\textwidth]{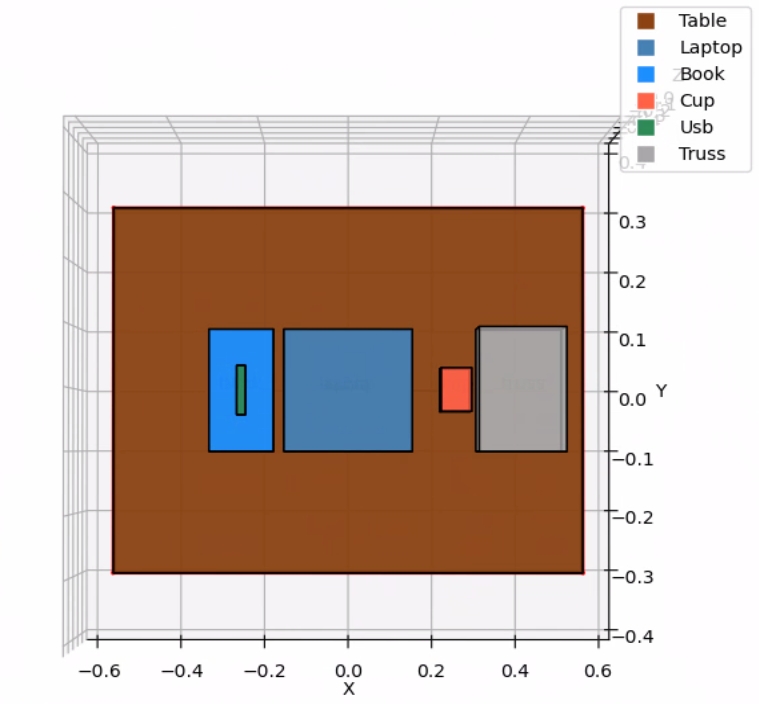} %
		\subcaption{Simulation Space}
	\end{minipage}
	\hspace{0.05\textwidth} %
	\begin{minipage}[b]{0.20\textwidth}
		\centering
		\includegraphics[width=\textwidth]{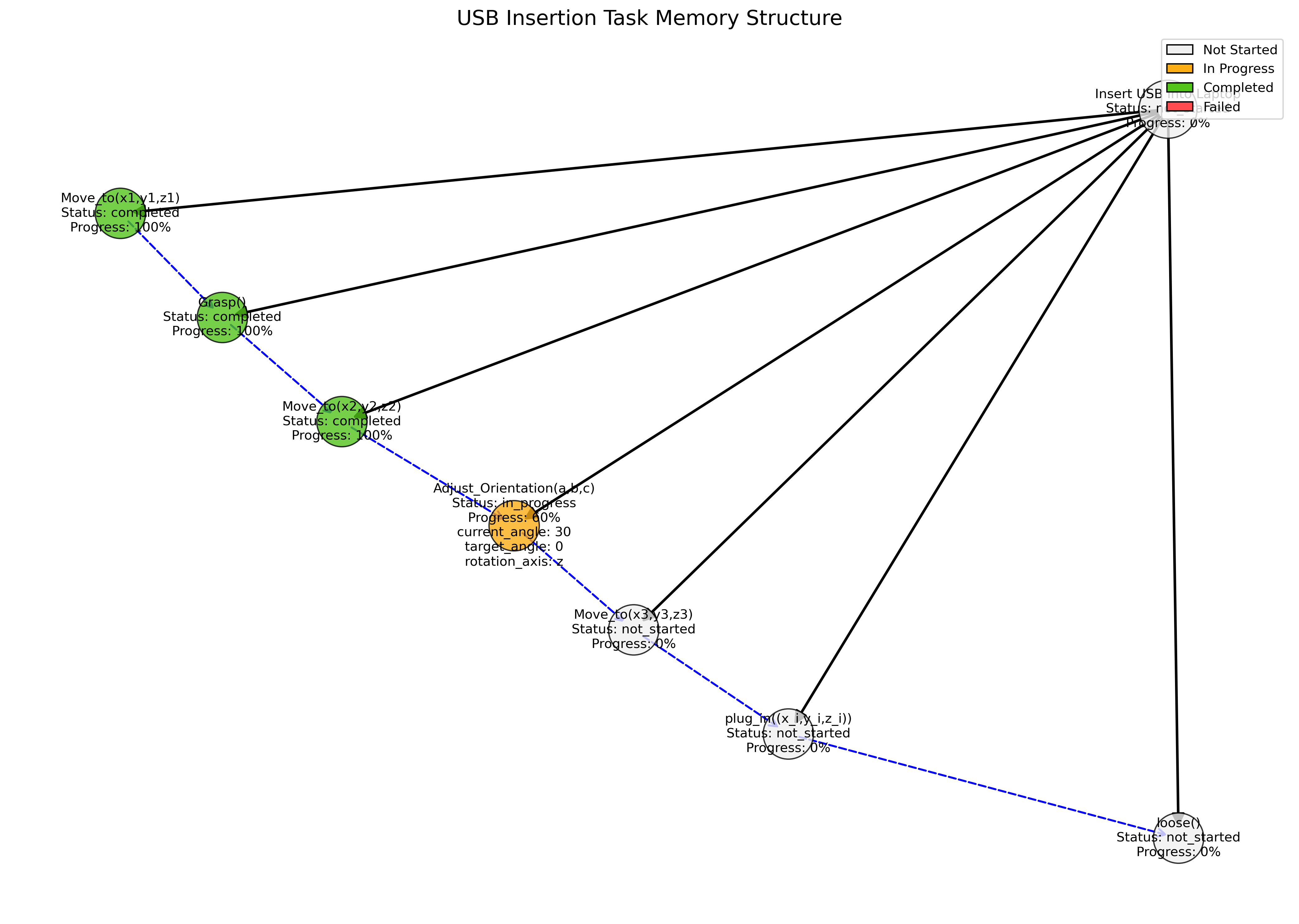} %
		\subcaption{Task Memory}
	\end{minipage}
	
	\vspace{0.05\textheight} %
	
	\begin{minipage}[b]{0.2\textwidth}
		\centering
		\includegraphics[width=\textwidth]{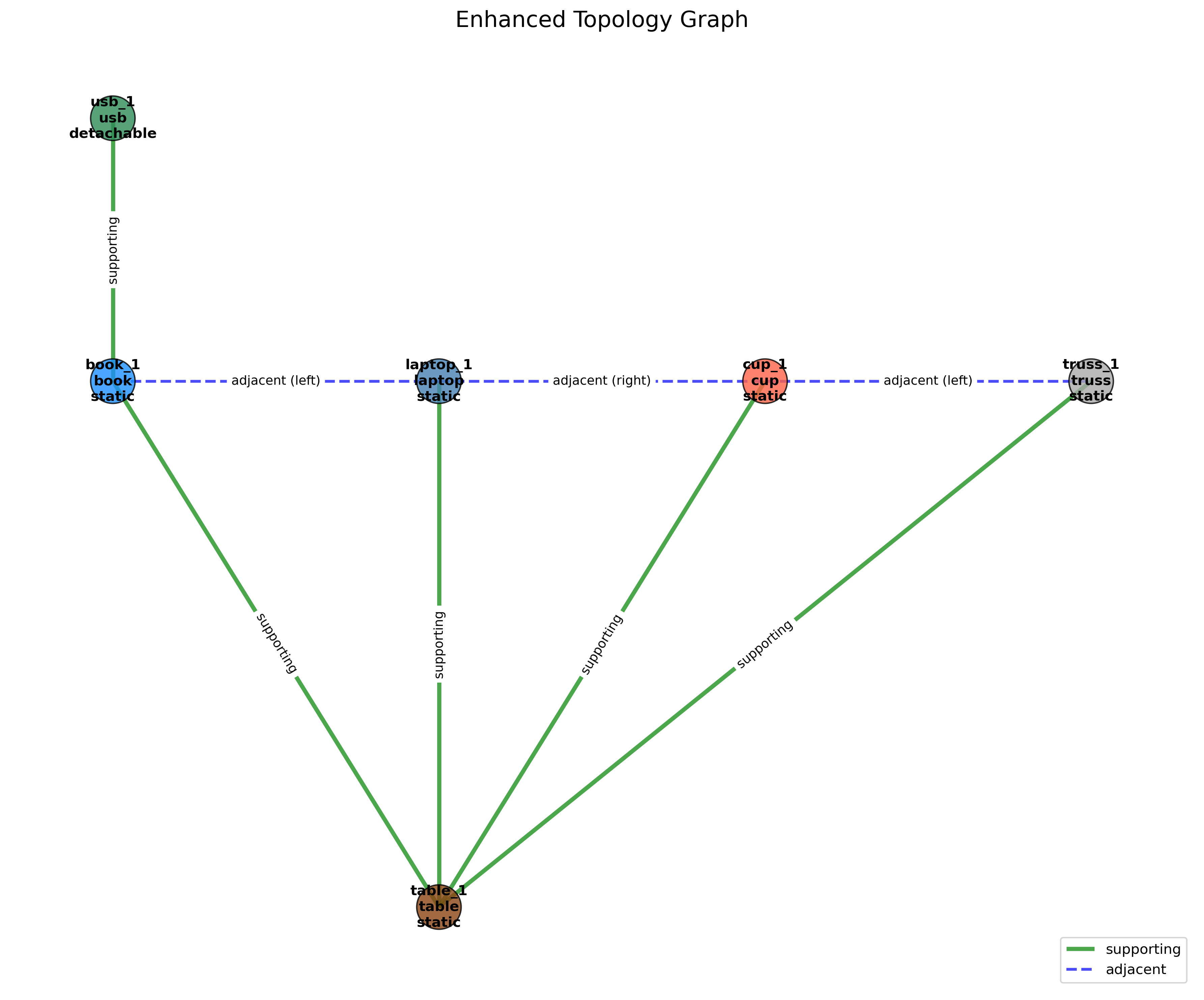} %
		\subcaption{Topology Graph}
	\end{minipage}
	\hspace{0.05\textwidth}
	\begin{minipage}[b]{0.2\textwidth}
		\centering
		\includegraphics[width=\textwidth]{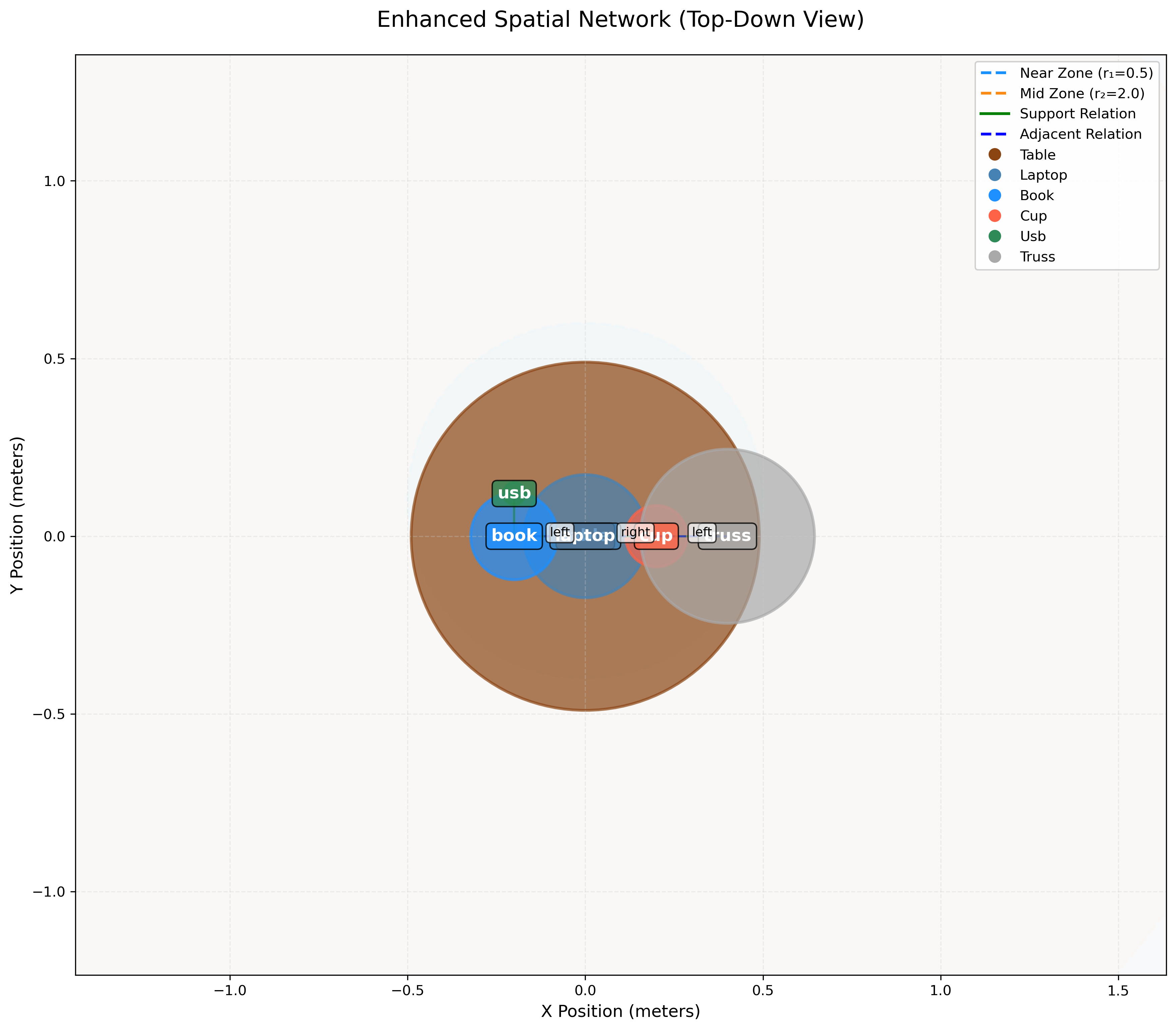} %
		\subcaption{Spatial Network}
	\end{minipage}
	
	\caption{Example of scene understanding and maintenance}
	\label{fig:four_images}
\end{figure}

Scene understanding in dynamic environments requires both semantic comprehension and precise geometric tracking. To address this, we propose a spatial understanding method that utilizes dual-layer representation. This method memorizes and provides both semantic and spatial information to the Vision-Language Model, thereby facilitating robot task planning and enabling precise manipulation.See figure.\ref{fig:four_images}.

\subsubsection{Dual-Layer Representation}
Our scene representation integrates semantic understanding with geometric information through a topology graph and spatial network structure.

\paragraph{2D Topology Graph} The topology graph \( G = (V, E, F) \) captures semantic relationships and object states.
\begin{itemize}
	\item Vertices
	\begin{equation}
		V = \{v_i \mid v_i = (\text{id}_i, \text{category}_i, \text{spatial\_index}_i, \text{state}_i)\}
	\end{equation}
	where \( \text{id}_i \) is object identifier, \( \text{category}_i \) represents its semantic class, \( \text{spatial\_index}_i \) links to Gaussian envelope in the spatial network, and \( \text{state}_i \) tracks the interaction state.
	
	\item Edges
	\begin{equation}
		E = \{e_{ij} \mid e_{ij} = (\text{type}_{ij}, \mathcal{R}_{ij})\}
	\end{equation}
	where \( \text{type}_{ij} \) represents the relationship type (e.g., "supporting", "adjacent") and \( \mathcal{R}_{ij} \) defines the physical constraints between objects.
	
	\item Feature Set
	\begin{equation}
		F = \{f_i \mid f_i = \text{JSON}(\text{VLM\_perception}_i)\}
	\end{equation}
\end{itemize}

These features are continuously updated through structured prompts requesting scene-relevant information, such as object properties, manipulation constraints, and task-specific attributes.

\paragraph{Spatial Network} The spatial network \( S \) maintains geometric information using minimal-volume Gaussian envelopes:
\begin{equation}
	S = \{s_i \mid i \in \text{spatial\_index}\}
\end{equation}

Each envelope \( s_i \) is parameterized by its center and covariance matrix:
\begin{equation}
	s_i = (\mu_i, \Sigma_i)
\end{equation}
where \(\mu_i \in \mathbb{R}^3\) represents the center position and \(\Sigma_i \in \mathbb{R}^{3\times3}\) defines spatial extent and orientation. The spatial\_index connects topology graph objects to their corresponding Gaussian envelopes, enabling integrated semantic-geometric representation.

\subsubsection{Topology Graph Update Strategy}
The topology graph update process integrates VLM-based semantic understanding with geometric verification to ensure accurate representation of object relationships.

\paragraph{Semantic Relationship Determination} 
The VLM analyzes the scene to determine semantic relationships between objects:
\begin{equation}
	\text{SR}(i,j) = \text{VLM}(F_i, F_j, \text{task\_description})
\end{equation}
where the VLM outputs structured relationship descriptions between object pairs.

\paragraph{Spatial Verification}
For each semantic relationship, we verify if the spatial configuration between Gaussian envelopes supports it. Given centers $\mu_i$, $\mu_j$ and covariance matrices $\Sigma_i$, $\Sigma_j$, we first compute the normalized distance:
\begin{equation}
	d_{ij} = \|\mu_i - \mu_j\|/\sqrt{\lambda_{\text{max}}}
\end{equation}
where $\lambda_{max}$ is the maximum eigenvalue of $\Sigma_i + \Sigma_j$, providing a conservative estimate of the combined spatial extent.

The geometric validation is then performed as:
\begin{align}
	&\text{ValidateGeometry}(s_i, s_j, \text{SR}(i,j)) = \notag \\
	&\begin{cases}
		1 & \text{if } d_{ij} \leq 2 \text{ and SR}(i,j) = \text{"containing"} \\
		1 & \text{if } 2 < d_{ij} \leq 3 \text{ and SR}(i,j) = \text{"contact"} \\
		1 & \text{if } 3 < d_{ij} \leq 6 \text{ and SR}(i,j) = \text{"nearby"} \\
		1 & \text{if } d_{ij} > 6 \text{ and SR}(i,j) = \text{"separate"} \\
		0 & \text{otherwise}
	\end{cases}
\end{align}
Here, the thresholds are determined based on probabilistic theory, where \( 3\sigma \) covers approximately 99\% of the probability mass in a Gaussian distribution, \( 2\sigma \) covers 94\%, and \( 6\sigma \) is determined empirically.

\paragraph{Edge Management}
Edges in the topology graph are established based on the validation result:
\begin{equation}
	 \text{AddEdge}_{ij} \iff \text{ValidateGeometry}(s_i, s_j, \text{SR}(i,j))
\end{equation}

This dual verification process ensures that the topology graph accurately reflects both semantic understanding and physical reality, providing a reliable foundation for subsequent task planning and execution.

\subsubsection{Spatial Network Management Strategy}
The spatial network employs an adaptive update strategy based on distance-aware processing zones, enabling efficient resource allocation while maintaining appropriate precision levels:
\begin{equation}
	\mathcal{Z}(d) = \begin{cases}
		\text{Far Zone} & d \geq r_2 \\
		\text{Mid Zone} & r_1 \leq d < r_2 \\
		\text{Near Zone} & d < r_1
	\end{cases}
\end{equation}

\paragraph{Gaussian Envelope Update Framework}
We represent each object using a minimal-volume Gaussian envelope that encloses all its observed points. Upon initial object detection, we place a unit Gaussian envelope at the 3D position corresponding to the object's centroid pixel, with mean \(\mu_i^0\) and \(\Sigma_i^0 = I\).

For each new point cloud observation, we determine valid points for updating envelope i through both semantic consistency and the Mahalanobis distance criterion:
\begin{equation}
	(p - \mu_i^{t-1})^T(\Sigma_i^{t-1})^{-1}(p - \mu_i^{t-1}) \leq \gamma
\end{equation}

The envelope update follows a two-step process. First, we update the center position through weighted averaging:
\begin{equation}
	\mu_i^t = \alpha\mu_i^{t-1} + (1-\alpha)\frac{1}{|\mathcal{P}_{new}|}\sum_{p \in \mathcal{P}_{new}} p
\end{equation}
where \(\alpha\) controls the temporal smoothing. Then, we optimize the covariance matrix to maintain minimal volume while enclosing all valid points:
\begin{equation}
	\begin{aligned}
		\{\mu_i^t, \Sigma_i^t\} &= \argmin_{\mu, \Sigma} \text{tr}(\Sigma) \\
		\text{s.t.} &\quad (p - \mu)^T\Sigma^{-1}(p - \mu) \leq 1, \forall p \in \mathcal{P}_i^t
	\end{aligned}
\end{equation}

\paragraph{Zone-Specific Processing Strategies}
We adapt our spatial network maintenance strategy based on distance from the interaction focus, balancing computational efficiency with geometric accuracy. In the far zone ($d \geq r_2$), we maintain a simplified representation using a standardized unit Gaussian envelope. The mean of this envelope is updated to the 3D position corresponding to the object's centroid in the latest image frame:
\begin{equation}
	\mu_i^t = \text{Pixel2Space}(c_i^t), \quad \Sigma_i^t = I
\end{equation}

For objects in the mid zone ($r_1 \leq d < r_2$), we employ the base optimization framework but with relaxed convergence criteria to balance accuracy and computational cost:
\begin{equation}
	\|\mu_i^t - \mu_i^{t-1}\| \leq \delta_{mid}, \quad \|\Sigma_i^t - \Sigma_i^{t-1}\|_F \leq \epsilon_{mid}
\end{equation}

In the near zone ($d < r_1$), where precise object interaction may be required, we implement comprehensive geometric tracking through high-precision envelope optimization:
\begin{equation}
	\|\Sigma_i^t - \Sigma_i^{t-1}\|_F \leq \epsilon_{max}
\end{equation}
Additionally, we maintain detailed component-level representations:
\begin{equation}
	V_i = \{\mathcal{V}_j \mid j \in \text{SubComponents}(i)\}
\end{equation}
where each component is discretized into a fine-grained voxel grid:
\begin{equation}
	\mathcal{V}_j = \text{Voxelize}(\{p \in \mathcal{P}_i^t \mid \text{SubLabel}(p) = j\}, 1\text{mm})
\end{equation}

\subsection{Task-Oriented VLM Interaction}
\label{task_oriented_vlm_interaction}
Building upon our dual-layer representation, we design a structured VLM interaction mechanism that dynamically generates prompts to guide robot actions. This mechanism hinges on two key components: our prompt engineering strategy and task memory maintenance system, which are elaborated in this section.

\subsubsection{Task Memory Structure}
Our system maintains a persistent memory structure $\mathcal{M}$ that decomposes the task into a topological path with execution states. It tracks subtask dependencies, progress, and historical task performance, enabling informed decision-making. The structure includes:

\begin{itemize}
	\item $TTP$: Hierarchical subtask dependencies and execution order, guiding task completion.
	\item $SS$: Real-time status indicators reflecting the current task progress.
	\item $MSH$: A stack of previously executed tasks, aiding in recognizing recurring patterns and informing future decisions.
\end{itemize}

\subsubsection{Dynamic Prompt Engineering}
We implement mode-specific prompt templates that combine scene information and task memory to guide robot actions. These prompts are dynamically generated based on the task's current state and the robot's historical performance. The system adapts its prompts according to the complexity of the task, switching between coarse and fine-grained instructions as required.
%

\paragraph{Coarse Motion Prompts}  
For coarse motion, prompts include:
\begin{equation}
	\begin{split}
		Prompt&_{coarse} = \\
		\Big[\, & TTP_{motion} \quad \text{: immediate manipulation goal,} \\
		& E_{motion} \quad \text{: topological relationships,} \\
		& SS_{motion} \quad \text{: subtask status,} \\
		& S_{global} \quad \text{: spatial constraints,} \\
		& MSH_{similar} \quad \text{: historical patterns} \,\Big]
	\end{split}
\end{equation}
These components maintain a high-level task overview while managing obstacles and constraints.

\paragraph{Fine Manipulation Prompts}  
For fine manipulation, we use more detailed prompts focusing on local geometry:
\begin{equation}
	\begin{split}
		Prompt&_{fine} = \\
		\Big[\, & TTP_{motion} \quad \text{: immediate manipulation goal,} \\
		& S_{local} \quad \text{: local geometric descriptions,} \\
		& D_{t} \quad \text{: real-time sensor feedback,} \\
		& SS_{motion} \quad \text{: subtask status,} \\
		& MSH_{similar} \quad \text{: prior manipulation patterns} \,\Big]
	\end{split}
\end{equation}
These prompts incorporate local geometry and real-time feedback to guide precise robot movements.

This adaptive prompting strategy allows the robot to efficiently allocate resources and adjust its actions according to task complexity, significantly improving task performance across various scenarios.

\section{Experiments}
We rigorously evaluated our progressive VLM planning system on precision-critical manipulation tasks. Our focus was on quantifying the impact of our dual-representation architecture and adaptive execution strategy rather than presenting incremental improvements over existing methods.

\begin{table*}[!htbp]
	\centering
	\caption{Cross-Task Performance Comparison with Full Metric Dimensions}
	\label{tab:performance_comparison}
	\resizebox{0.9\textwidth}{!}{%
		\begin{tabular}{llcccccccc}
			\toprule
			& & \multicolumn{4}{c}{\textbf{Task 1: Truss Assembly}} & \multicolumn{4}{c}{\textbf{Task 2: Connector Docking}} \\
			\cmidrule(lr){3-6} \cmidrule(lr){7-10}
			\textbf{Type} & \textbf{Model} & \textbf{SLPC} & \textbf{TPSR} & \textbf{MSR} & \textbf{TSR} & \textbf{SLPC} & \textbf{TPSR} & \textbf{MSR} & \textbf{TSR} \\
			\midrule
			LLM+Text & DeepSeek-R1 & / & 56.44\% & 89.40\% & 14.19\% & / & 32.67\% & 86.36\% & 18.34\% \\
			\midrule
			\multirow{1}{*}{Pure VLM}
			& GPT-4o & 0\% & 88.12\% & 25.26\% & 0\% & 0\% & 86.14\% & 25.83\% & 0\% \\
			\midrule
			\multirow{2}{*}{VLM+Ours}
			& Claude 3.5 & 61.67\%  & 94.16\% & 99.01\% & 91.01\% & 100\% & 88.20\% & 95.64\% & 81.21\% \\
			& MiniCPMv2.6 & 73.33\% & 90.41\% & 98.42\% & 82.75\% & 100\% & 91.96\% & 94.26\% & 74.96\% \\
			\midrule
			MMLM(FTed) & RDT-170M & / & / & 89.41\% & 57.14\% & / & / & 89.11\% & 50.07\% \\
			\bottomrule
		\end{tabular}%
	}
\end{table*}

\subsection{Experimental Setup}
Experiments were conducted using a 7-DOF Franka arm with a parallel gripper and wrist-mounted Intel RealSense RGB-D camera. Computations were performed on an NVIDIA RTX 4090D GPU (24GB VRAM). We evaluated five VLM-based models with task planning capabilities: Deepseek-r1, GPT4o, Claude 3.5, MiniCPMv2.6, and RDT1 (finetuned with task-specific data). The baseline comparison focused on assessing the added value of our progressive planning architecture for enhancing VLM-based manipulation.

\subsection{Benchmark Tasks}
We selected two benchmark tasks to evaluate the core capabilities of our VLM-based system: task switching between coarse and fine manipulation, and maintaining accuracy in dynamic environments. These tasks are highly relevant to industrial applications requiring precision and adaptability.

\subsubsection{Truss Assembly}
This task tested semantic understanding and spatial reasoning in cluttered environments. The setup included two target truss components, funnel-like slots for alignment, and distractor objects. The main challenge was identifying task-relevant objects and understanding their spatial relationships. Performance was measured by the SLPC metric, which evaluates scene-wide semantic-location pairing and coherent task planning in complex scenes.

\subsubsection{Aviation Connector Docking}
This task focused on high-precision manipulation and error recovery. Aviation connectors require precise grasp pose identification, sub-millimeter alignment during insertion, and real-time error detection/recovery upon failure. The evaluation emphasized precision control and adaptive strategies for fine manipulation.

\subsection{Results and Analysis}

\subsubsection{Evaluation Matrix}
\begin{itemize}  
	\footnotesize 
	
	\item \textbf{SLPC} (Semantic-Location Pair Correctness): Measures the accuracy of identifying and locating task-relevant objects.
	\newline
	\hspace*{\labelsep}  $SLPC = \frac{\text{Number of Correct Semantic-Location Pairs}}{\text{Total Ground Truth Pairs}} \times 100\%$
	
	\item \textbf{TPSR} (Task Planning Success Rate):  Evaluates the success rate of generating valid task plans.
	\newline
	\hspace*{\labelsep}  $TPSR = \frac{\text{Number of Successful Task Plans}}{\text{Total Task Planning Attempts}} \times 100\%$
	
	\item \textbf{MSR} (Motion Success Rate):  Measures the success rate of executing individual motion sub-tasks.
	\newline
	\hspace*{\labelsep}  $MSR = \frac{\text{Number of Successful Subtask Executions}}{\text{Total Subtask Execution Attempts}} \times 100\%$
	
	\item \textbf{TSR} (Task Success Rate):  Indicates the overall success rate of completing the entire task.
	\newline
	\hspace*{\labelsep}  $TSR = \frac{\text{Number of Successful Tasks}}{\text{Total Task Attempts}} \times 100\%$
\end{itemize}

\subsubsection{Performance Comparison}

The results from 200+ planning outcomes per configuration, presented in Table \ref{tab:performance_comparison}, demonstrate dramatic performance improvements with our progressive framework. 

Text-only LLM approaches (DeepSeek-R1) achieve only 56.44\% TPSR for truss assembly, lower than VLM models, confirming that visual inputs enhance spatial reasoning ability. However, pure VLM approaches (GPT-4o) show a critical disconnect: despite high planning success (88.12\% TPSR), they completely fail at task execution (0\% TSR) due to poor motion success rates (25.26\% MSR).

In contrast, our framework achieves up to 91.01\% TSR for truss assembly and 81.21\% TSR for connector docking. This stark improvement confirms two key contributions of our approach:

\paragraph{Enhanced 3D spatial understanding} The complete failure of semantic-location pairing in pure VLM approaches (0\% SLPC) versus high values with our framework (61.67\%-100\%) demonstrates our architecture's superior grounding of visual-language reasoning in physical space.

\paragraph{Bridging planning-execution gap} Our framework maintains high motion success rates (94\%) across both tasks, directly addressing the critical limitation of pure VLMs that cannot translate plans into successful physical actions.

Despite our efforts to finetune RDT-170M with task-specific demonstration data, our framework still outperforms it by substantial margins (34-37\% higher TSR). This confirms that our architectural innovations address fundamental limitations that cannot be overcome through additional training alone.

\begin{table}[h!]
	\centering
	\caption{Ablation Study (using MiniCPM v2.6) in Connector Docking Task}
	\label{tab:ablation}
	\begin{tabular}{lcccc}
		\hline
		\textbf{Configuration}    & \textbf{SLPC} & \textbf{TPSR}    & \textbf{MSR}    & \textbf{TSR}  \\
		\hline
		All systems               & 100\%         & 91.96\%          & 94.26\%          & 74.96\%      \\
		w/o Task Memory           & 100\%         & \textbf{83.60\%} & 94.26\%          & 68.21\%      \\
		w/o Dual-Layer Module     & \textbf{0\%}  & 91.96\%          & 24.09\%          & 0\%          \\
		w/o Interaction Algorithm & 100\%         & 91.96\%          & \textbf{72.28\%} & 51.96\%      \\
		\hline
	\end{tabular}
\end{table}

\subsubsection{Ablation Analysis}
To isolate the contribution of key components, we performed targeted ablation studies using miniCPM v2.6, shown in Table \ref{tab:ablation}.

\subsubsection{Ablation Analysis}
Our ablation studies, presented in Table \ref{tab:ablation}, reveal three critical insights about our system's components:

\paragraph{Task memory structure} It provides a 6.75\% improvement to overall TSR by enhancing planning quality (TPSR improved from 83.60\% to 91.96\%). This improvement stems from the system's ability to identify previously challenging subtasks and apply successful historical strategies to these cases, significantly boosting success rates for traditionally difficult manipulation sequences.

\paragraph{Dual-layer module} This module is fundamental to task completion. Without it, SLPC drops to 0\% and TSR to 0\%, despite maintaining high planning success (91.96\% TPSR). This dramatic failure occurs because the system loses all ability to ground semantic concepts in physical space, confirming that our dual representation enables the essential bridge between high-level understanding and precise spatial configurations.

\paragraph{Adaptive interaction algorithm} This algorithm contributes a substantial 23\% improvement to TSR (74.96\% vs 51.96\%). Most notably, it significantly enhances MSR (94.26\% vs 72.28\%) by implementing repeated fine manipulation attempts with continuous visual feedback after each attempt. This iterative observation-execution cycle is the key mechanism that enables precise motion adjustments until successful completion is achieved.

\subsubsection{Broader Implications}
Our experimental results reveal several significant implications beyond the immediate performance metrics:
\paragraph{Modular enhancement for any VLM} Our framework provides a universal enhancement layer that integrates with any VLM without architectural modifications. This modularity enables robotic systems to leverage future VLM advancements while maintaining reliable physical grounding.
\paragraph{Cloud-to-edge deployment flexibility} By utilizing API-based VLM interaction, our approach decouples sophisticated reasoning from onboard computing, enabling deployment on resource-constrained robotic platforms while maintaining high performance.
\paragraph{Reduced data requirements} Unlike fine-tuned approaches that require extensive demonstration data yet still underperform (as seen with RDT-170M), our structural approach addresses core limitations directly, suggesting a more efficient path toward generalizable robotic manipulation.
\paragraph{Resilience against cognitive limitations} Our topology graph, spatial network, and task memory structures effectively mitigate hallucination and working memory limitations inherent to current VLMs and LLMs—particularly crucial for extended, complex manipulation sequences.


\section{CONCLUSIONS}
Our progressive VLM planning framework successfully bridges the gap between high-level vision-language understanding and precise robotic manipulation. Experimental results demonstrate that our approach achieves up to 91.01\% success rates on precision-critical tasks, where pure VLM methods fail completely. The experiments confirm the importance of our dual representation and adaptive execution strategy, with performance degrading significantly when either component is removed. These findings validate our core hypothesis that effective robotic VLM applications require both semantic grounding in physical space and dynamic resource allocation between coarse and fine manipulation phases. This modular approach provides a pathway for leveraging increasingly powerful VLMs in robotics, while ensuring deployment flexibility across diverse computing environments.

\addtolength{\textheight}{-12cm}   



%
%
%
%
%
%

\bibliographystyle{IEEEtran}
\bibliography{reference}

\end{document}